\documentclass[10pt,twocolumn,letterpaper]{article}

\usepackage[pagenumbers]{cvpr} 

\usepackage[dvipsnames]{xcolor}

\usepackage{marvosym}

\definecolor{cvprblue}{rgb}{0.21,0.49,0.74}
\usepackage[pagebackref,breaklinks,colorlinks,citecolor=cvprblue]{hyperref}
\usepackage{colortbl}

\definecolor{mygray}{gray}{.92}

\usepackage{newfloat}
\usepackage{booktabs} 

\def\paperID{14242} 
\def\confName{CVPR}
\def\confYear{2024}

\title{Smart Help: Strategic Opponent Modeling for \\ Proactive and Adaptive Robot Assistance in Households}

\author{
Zhihao Cao$^{1,2,}\thanks{Indicates equal contribution. Work was conducted during Zhihao Cao and Zidong Wang's internships at BIGAI. \textrm{\Letter} Corresponding author: Lifeng Fan (lifengfan@bigai.ai). Our environment, dataset, and codes are available at \url{https://github.com/caozh20/SmartHelp}.}$~,\hspace{1pt}
Zidong Wang$^{1,2,*}$,\hspace{1pt}
Siwen Xie$^{3}$,\hspace{1pt} 
Anji Liu$^{4}$,\hspace{1pt}
Lifeng Fan$^{1,\textrm{\Letter}}$\\
\small $^1$ State Key Laboratory of General Artificial Intelligence, Beijing Institute for General Artificial Intelligence (BIGAI)\\
\small $^2$ Department of Automation, Tsinghua University
\small $^3$ Yuanpei College, Peking University\quad{} \\
\small $^4$ Computer Science Department, University of California, Los Angeles\\
\vspace{-9pt}
}

\begin{document}
\maketitle

\begin{abstract}

Despite the significant demand for assistive technology among vulnerable groups (e.g., the elderly, children, and the disabled) in daily tasks, research into advanced AI-driven assistive solutions that genuinely accommodate their diverse needs remains sparse. Traditional human-machine interaction tasks often require machines to simply help without nuanced consideration of human abilities and feelings, such as their opportunity for practice and learning, sense of self-improvement, and self-esteem. Addressing this gap, we define a pivotal and novel challenge \textit{Smart Help}, which aims to provide proactive yet adaptive support to human agents with diverse disabilities and dynamic goals in various tasks and environments. To establish this challenge, we leverage AI2-THOR~\cite{kolve2017ai2} to build a new interactive 3D realistic household environment for the \textit{Smart Help} task. We introduce an innovative opponent modeling module that provides a nuanced understanding of the main agent's capabilities and goals, in order to optimize the assisting agent's helping policy. Rigorous experiments validate the efficacy of our model components and show the superiority of our holistic approach against established baselines. Our findings illustrate the potential of AI-imbued assistive robots in improving the well-being of vulnerable groups. 

\end{abstract}  

\section{Introduction}
\label{sec:intro}

\begin{figure*}[h]
\vspace{-0.3cm}
   \includegraphics[width=\textwidth]{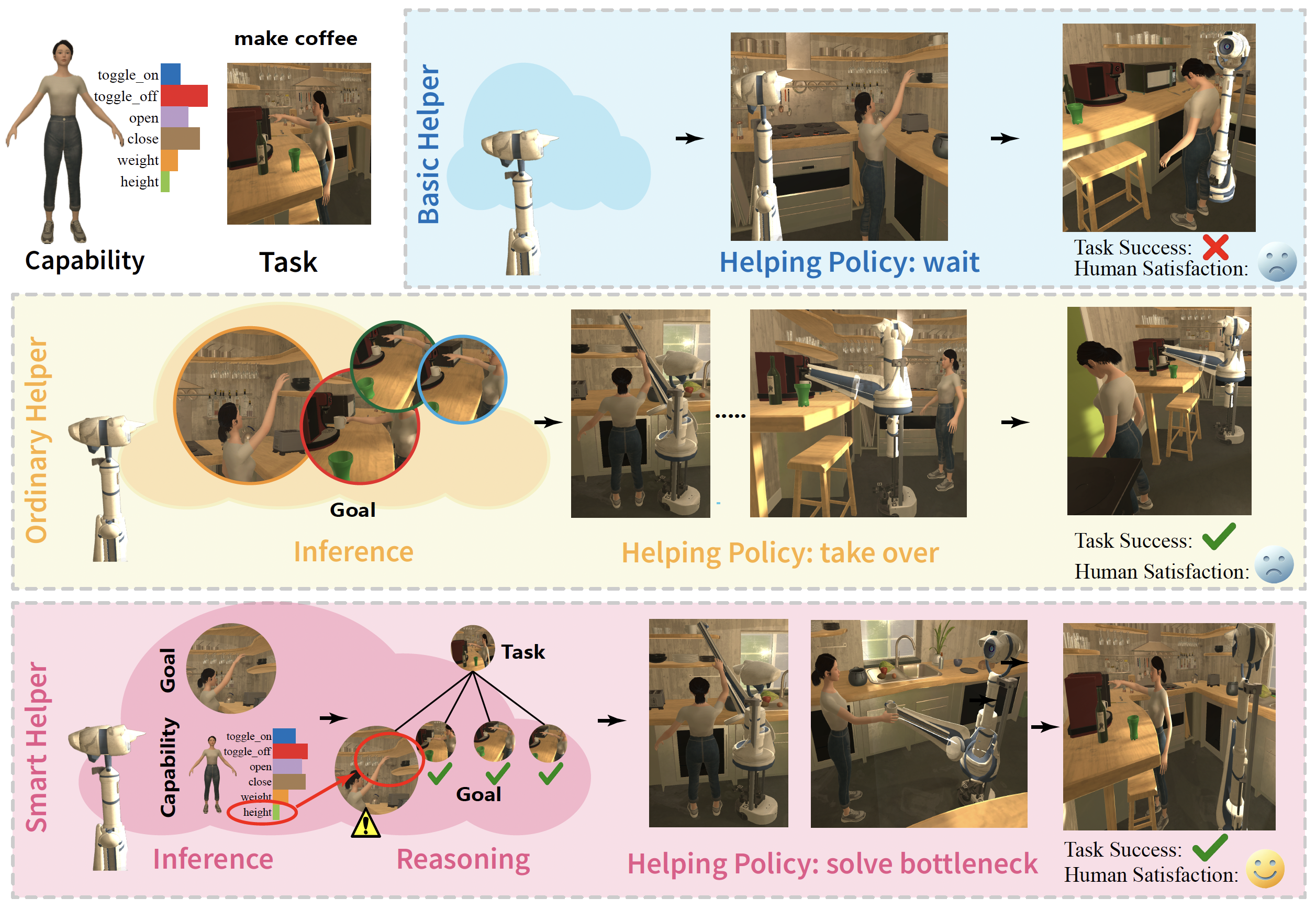}
\vspace{-0.85cm}
   \caption{ \textbf{An example of the \textit{Smart Help} strategy for an assistive robot}. The top left shows the human user's capability distribution across several dimensions (e.g., \textit{weight} for lifting heavy objects and \textit{height} for grasping objects from high positions), as well as the current task (e.g., making coffee). The figure illustrates three types of helpers: (1) The \textit{ \textbf{Basic Helper}} merely observes the human user without inferring anything, and thus remains idle and waits until the human user fails the task, leading to frustration. (2) The \textit{\textbf{Ordinary Helper}} infers the human user's goals through observations and always provides simple and direct assistance for each goal, taking over the entire task. While the task is successfully completed, the human user is left dissatisfied because the robot's helping strategy is overly intrusive, causing discomfort. (3) The \textit{\textbf{ Smart Helper}} infers both the human user's goals and her capability distribution. It reasons about whether the human user is capable of independently accomplishing each goal and thus identifies the bottleneck goal that hinders task completion. The proactive helper then adapts its strategy to assist solely with the bottleneck goal. As a result, the human user successfully completes the task with tailored support, feeling comfortable and satisfied. 
   }
\label{fig:teaser}
\vspace{-0.45cm}
\end{figure*}

All of us may find ourselves within vulnerable demographics at some point. Throughout the human life span, we confront an array of challenges, whether originating from physical discomfort~\cite{davila2019physical}, emotional turmoil~\cite{van2021unpacking}, or the inevitable march of aging~\cite{harman2001aging}, that can hinder our ability to perform even the simplest tasks that we once accomplished effortlessly (e.g., lifting a heavy object). This predicament often triggers a complex emotional response. The fear of being labeled as ``special" or ``disabled" can evoke feelings of diminished self-esteem, reduced self-efficacy and self-sufficiency, and a sense of personal boundary violation~\cite{buoncompagni2017towards, parida2021pandemic}. However, previous research in human-robot interaction has primarily focused on pure cooperation~\cite{jiang2018learning, xuan2001communication, panait2005cooperative, puig2020watch, puig2023nopa, ullman2009help}, and assistive technologies~\cite{bemelmans2012socially, brose2010role} have been designed to simply take over everything for human users, often disregarding their emotional well-being. Considering the importance of sensitivity and consideration when offering help, we need new assistance technologies with an emphasis not only on the successful completion of the task but also on the recipient's emotional acceptance of the assistance~\cite{buoncompagni2017towards, parida2021pandemic}. This introduces a new dimension to the concept of \textit{learning to help} in AI, extending the boundaries beyond technical proficiency and into the realm of empathetic engagement. 

To tackle this issue, we propose a novel and pivotal challenge \textit{Smart Help}, aiming to provide both proactive and adaptive support to human agents with diverse disabilities and dynamic goals in different tasks and environments. \cref{fig:teaser} exemplifies the concept of the \textit{Smart Help} strategy as applied to an assistive robot aiding a human in a household task. The \textit{Basic Helper} fails to provide proactive assistance due to its inability to infer the user's goals accurately. The \textit{Ordinary Helper} successfully infers the user's goals from observations and offers proactive help, but it lacks adaptability to different user needs. Its simple ``take-over'' helping strategy can cause a feeling of discomfort in the human user. In contrast, the \textit{Smart Helper} not only infers both the user's goals distribution but also reasons about the user's capability of completing each goal independently. As a result, the \textit{Smart Helper} can identify critical goals that hinder task completion and provide proactive and adaptive help that selectively addresses those bottlenecks. This approach allows the user to accomplish the task successfully and joyfully.

Leveraging AI2-THOR~\cite{kolve2017ai2}, a 3D interactive environment engineered for realistic home simulations, we demonstrate a concrete realization of the proposed novel challenge \textit{Smart Help}.  
We create a multi-agent interactive environment featuring a main agent with diverse capability distributions (representing the vulnerable group) across dimensions, such as toggling, opening, closing, weight, and height. 
Within this environment, the main agent faces challenges in achieving dynamic goals across various tasks, and an assistive robot is introduced to assist the main agent throughout their endeavors. 
Unlike previous assistive tasks that were highly specialized and constrained, such as the use of a robotic arm for the disabled~\cite{casey2021bci, uehara2010mobile, fall2015intuitive}, our work represents, to the best of our knowledge, the first construction of a 3D home environment designed to assist the vulnerable group with various general daily household tasks.

Furthermore, we build a new benchmark model for our proposed challenge, which consists of 1) an opponent modeling module that jointly optimizes goal inference and capability estimation, and 2) a helping policy module that reasons about the bottleneck, and learns an optimal \textit{Smart Help} policy in an online interactive manner. To better evaluate performance in the \textit{Smart Help} task, we also introduce six assessment metrics. Contrary to traditional evaluation metrics primarily focusing on cumulative reward or task completion, our metrics emphasize the helper's contribution to the task and the essentiality of assistance. Rigorous experiments validate the efficacy of our model components and show the superiority of our holistic approach against baselines. We believe our proposed task, environment, model, and benchmark will contribute to the development of next-generation advanced home assistive robot technologies.

Our \textbf{main contributions} can be summarized as follows:
\begin{itemize}
    \item We propose a novel \textit{Smart Help} challenge that aims at learning to provide both proactive and adaptive help to diverse human users (especially vulnerable groups) based on inferred goals and capabilities.
    \item To the best of our knowledge, we contribute the first 3D realistic home environment built upon AI2-THOR, that focuses on assisting the vulnerable group with daily household tasks in an online and interactive manner. 
    \item We provide a benchmark model with a joint goal and capability inference, bottleneck reasoning, and helping policy improvement. Strict experiments and the proposed holistic evaluations validate the efficacy of our model.
\end{itemize}

\section{Related Work}

\noindent \textbf{Assistive Robots.}
The domain of assistive robots encompasses a broad spectrum of research, focusing on enhancing the life quality for individuals with various needs and spanning diverse applications such as mobility aids \citep{aigner1999shared, glover2003robotically}, companion robots \citep{roy2000towards}, and robotic arms \citep{graf2002robotic}. \citet{feil2005defining} underscore that Socially Assistive Robotics entails robots assisting humans through effective interaction in tasks such as food delivery \citep{simmons1997layered}, healthcare \citep{inoue2008effective}, and other tasks necessitating social interaction. However, many of them only rely on simple rules or programs to achieve social interaction \cite{simmons1997layered, inoue2008effective}. The deployment of assistive robots has also raised ethical and safety considerations, particularly in terms of user autonomy, privacy, acceptance, trust, etc. \cite{broadbent2009acceptance, broadbent2010attitudes, feil2005defining, vcaic2018service}. The objective of our work is to devise an effective algorithm for Socially Assistive Robotics to estimate people's goals and capabilities, thereby enabling comfortable and smart assistance.


\noindent \textbf{Embodied Multi-agent Collaboration.} A series of embodied collaboration tasks have been developed recently thanks to the development of Embodied AI simulators~\cite{kolve2017ai2,Deitke2020roboTHOR,ehsani2021manipulaTHOR,puig2018virtualhome,li2021igibson,shen2021igibson,savva2019habitat1,szot2021habitat2}. However, these tasks always focus on some limited task completion, such as collaborative navigation~\cite{wang2021collaborative,liu2022multi} and collaborative furniture rearrangement~\cite{jain2019two,jain2020cordial}. Watch-and-Help~\cite{puig2020watch} and NOPA~\cite{puig2023nopa} study the cooperative  tasks between two agents with the same capabilities. Our work focuses on the collaboration between agents with different capabilities and study the strategic assistance for the vulnerable group on household tasks.

\noindent \textbf{Opponent Modeling.} Opponent modeling~\cite{opponent-model-he2016opponent, opponent-moel-foerster2017learning, opponent-model-everett2018learning}, a pivotal approach in multi-agent interaction tasks, leverages a variety of methodologies such as Inverse Reinforcement Learning~\citep{ohnbar2018personalized}, Bayesian methods~\citep{ullman2009help, rosman2016bayesian, hernandez2016bayesian, hernandez2017learning, zheng2018deep}, Deep Q-Networks (DQN)~\citep{he2016opponent, mnih2015human}, Variational Auto-Encoders~\citep{papoudakis2020variational}, Markov Decision Process~\cite{ohnbar2018personalized}, etc. This technique is especially valuable in competitive settings like games and strategic decision-making, where understanding and anticipating an opponent's behavior or unseen traits (e.g., Theory of Mind modeling~\cite{rabinowitz2018machine}) can significantly influence the outcome. It enables agents to predict future environmental state transitions and refine their strategies accordingly~\cite{opponent-model-rollout-zhang2021model, opponent-model-rollout-yu2022model}. Projects like Watch-and-Help~\cite{puig2020watch} and NOPA~\cite{puig2023nopa} have explored estimating a main agent's goals to improve coordination. Yet, these initiatives often overlook the capabilities of the main agent, leading to a gap in developing effective support strategies for those in need. Our research addresses this gap by focusing on enhancing the acceptance of AI assistance among users, as emphasized in~\cite{vcaic2018service}, highlighting the importance of aligning AI functionalities with user needs and preferences. 
Additionally, we propose to model the capability of the opponent and learn a smart, adaptive, and empathetic helping policy for vulnerable people in a challenging 3D household interactive environment with partial observation and high uncertainty. 

\section{The Smart Help Challenge}

\subsection{Problem Formulation}

We model the interaction between the main agent and the helper with a multi-agent Partially Observable Markov Decision Process (POMDP\citep{cassandra1994acting}), which is formally defined as a tuple $G = \textless S, A, O, R, T, n, \gamma \textgreater$. $S$ represents the state space, including physical states and mental states. $A$ is the joint action space for $n$ agents, whose local observations compose observation space $O$. $T(s' | s, a)$ denotes the transition probability. $R(s,a)$ denotes the shared reward function and $\gamma\in [0,1)$ is the discount factor. Specifically, there are $n=2$ agents in our \textit{Smart Help} Challenge, i.e., a main agent and a helper agent. The objective is to train a helper agent to assist the main agent in achieving goals, considering the emotional state of the main agent simultaneously. 

In previous work, the reward for the helper agent is:
\begin{equation}
    R_h(s,a_h) = R^{g_h}(s, a_h) + \beta R^{g_m}(s, a_h), \\
    \label{eq:reward_normal}
\end{equation}
where the $\beta \in (0, 1)$ is a factor controlling the level of \textit{altruism} of the helper. 
$R^{g_h}(s, a_h)$ is the reward attributed to the successful completion of the helper agent's goal at state $s$ through action $a_h$.
$R^{g_m}(s, a_h)$ is the same for the main agent's goal achievement.
This formulation assumes equal consideration for both agents when rewarding goal completion. However, when the helper fulfills a goal of the main agent, it triggers emotional responses in the main agent. Integrating this emotional component with ~\cref{eq:reward_normal}, we have:
\begin{equation}
    \begin{aligned}
        R_h(s,a_h) & = R^{g_h}(s, a_h) \\
        &+ \beta  (R^{g_m}(s, a_h) + \lambda_e R^{e_m}(s, a_h)), 
    \end{aligned}
    \label{eq:reward_emotion}
\end{equation}
where $R^{e_m}(s, a_h)$ means the reward correlated to the emotions of the main agent when the helper does the action $a_h$, and $\lambda_e$ is a hyper-parameter that controls the helper's sensitivity to the emotional states of the main agent. In our \textit{Smart Help} challenge, the helper agent's role is solely to assist the main agent in accomplishing his/her goals, without its own goals. So we omit the term $R^{g_h}(s, a_h)$ and set $\beta=1$, leading to the helper's reward as:
\begin{equation}
    R_h(s, a_h) = R^{g_m}(s, a_h) + \lambda_e R^{e_m}(s, a_h). \\
    \label{eq:reward_new}
\end{equation}
Indeed, this is what sets \textit{smart Help} apart from other assistance-related tasks. Through such design, we emphasize that AI agents must possess the capacity to discern not just how to aid humans, but also when their assistance is truly needed and will be valued.


\subsection{Challenge Implementation}

\begin{figure*}[t]
\begin{center}
   \includegraphics[width=1.0\textwidth]{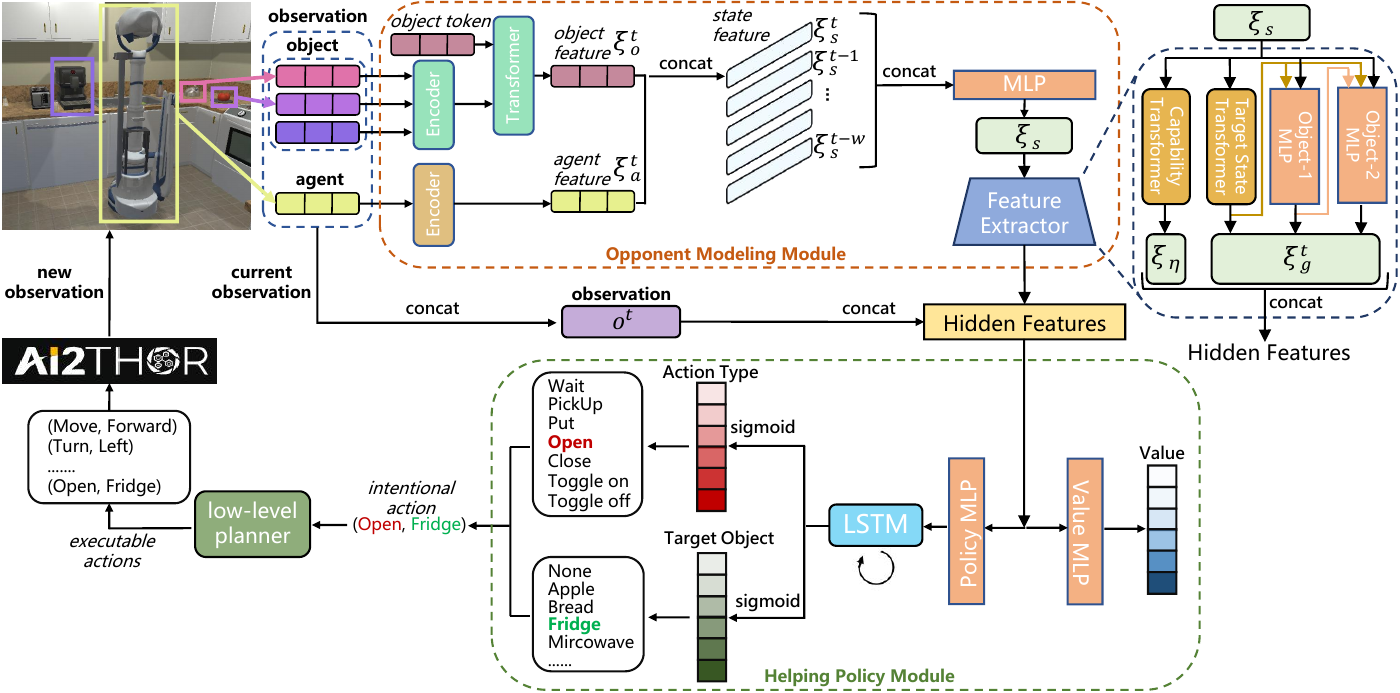}
\end{center}
\vspace{-0.5cm}
   \caption{
    This figure depicts the architecture of our \textit{smart help} model and its interaction mechanisms with the environment. 
    The model is partitioned into two primary components: an opponent modeling module and a helping policy module. 
    The opponent modeling module is designed to estimate the goal and capability of the main agent. 
    It adopts a sliding window paradigm and utilizes several MLP layers to generate the state feature $\xi_s$. This feature is then processed by a feature extractor to derive the capability feature $\xi_{\eta}$ and the goal feature $\xi^t_g$.
    The helping policy module is composed of two MLP layers to produce action decision and value estimation respectively. During interaction with the environment, the helping model outputs an intentional action, representing the target state it aims to reach. 
    The environment employs a low-level planner to decompose the intentional action into a sequence of executable actions, which are then processed by the AI2THOR simulator, guiding the assistant toward the intended state. 
    Finally, the simulator provides a new observation as feedback, triggering the next cycle of interaction.}
\label{fig:model}
\vspace{-0.3cm}
\end{figure*}

We implement a tangible version of \textit{Smart Help} in the AI2THOR~\cite{kolve2017ai2} simulator, which emphasizes adaptive helping policy for individuals with various disabilities, as shown in ~\cref{fig:teaser}. Our new environment includes a main agent that simulates human behavior in various assistive household tasks. The main agent, assigned with a new task and a capability distribution at the beginning of each round, attempts to complete the task with an expert planning policy, which utilizes complete information to divide the task into distinct goals and plan intentional actions for the goals. We also implement a low-level planner that translates the intentional action to a sequence of executable actions in the simulator, so as to enable the main agent to reach the target state indicated by the input intentional action. The workflow of the low-level planner is a loop involving the following steps:
\begin{enumerate}
    \item Navigate to the relevant object by following the shortest path determined by the object's position and the room layout; or be directly teleported to the target position.
    \item Interact with the object in accordance with the goal. 
\end{enumerate}
However, actions taken by the main agent might fail due to certain disabilities or the physical constraints of the scene. Thus, the helper agent, as in ~\cref{fig:model}, based on its symbolic observations of the world, should infer both the goal and capability of the main agent and provide proactive assistance with the bottleneck for the main agent. We will introduce the detailed environment settings in the following part. 

\noindent \textbf{Object State.}
We define object state of object $i$ as $e_i = (type_i, pos_i, w_i, attr_i, pr_i)$, where:
\begin{itemize}
\item $type_i \in \mathbb{N}$ denotes the type of object $i$ with an index;
\item $pos_i \in \mathbb{R}^3$ records the position of object $i$'s center;
\item $w_i \in \mathbb{R}$ represents the weight of object $i$;
\item $attr_i \in \{0, 1\}^{4+2}$ denotes the attributes of object $i$, including its 4 status (i.e., whether it is picked up, opened, cooked, and toggled on) and its visibility to the 2 agents;
\item $pr_i \in \mathbb{R}$ signifies the type of the parent receptacle of the object $i$ through an index; here, the parent receptacle means an object that serves as the container or the support for another subordinate object.
\end{itemize}

\noindent \textbf{Action Space.}
We employ intentional actions to construct the action space, where intentional actions represent the target state that the agent wants to achieve. It is defined as $a_i = (predicate_i, e_i) \in A$; for instance, (PickUp, 3), where 3 is an object index, representing ``Bread'' in our environment. 
There are seven kinds of different intentional actions: (Wait), (PickUp, $e_i$), (Put, $pr_i$), (ToggleOn, $e_i$), (ToggleOff, $e_i$), (Open, $e_i$), and (Close, $e_i$). 
In total, there are $| A | = 373$ possible intentional actions. 

\noindent \textbf{Agent Capability.}
The variance of agent capability is manifested in the differences in the transition matrix, which means that different agents will produce different outcomes for the same action at the same state. For example, some agents may be able to pick an object up, while others cannot due to the its weight. 
Learning the matrix itself is impractical due to the vast size of the space. Instead, we select six parameters to represent the distinct characteristics of different matrices. Therefore, the capability of agent $i$ is represented as $\eta_i=(\alpha_i, \beta_i, \gamma_i, \delta_i, \epsilon_i, \zeta_i)$ where:
\begin{itemize}
\item The maximum height an agent can reach $\alpha_i \in [0, 1]$ and the maximum weight an agent can lift $\beta_i \in [0, 1]$ determine whether an agent can execute the action \textit{PickUp}; 
\item $\gamma_i, \delta_i, \epsilon_i, \zeta_i \in [0, 1]$ respectively represent the agent's ability to complete the actions \textit{Open}, \textit{Close}, \textit{ToggleOn} and \textit{ToggleOff}. 
\end{itemize}

\noindent \textbf{Agent State.}
We define the state of agent $i$ as $h_i = (\eta_i, pos_i, rot_i, e_i, a_i, succ_i)$, wherein:
\begin{itemize}
\item $\eta_i \in C = [0, 1]^6$ denotes the capability of agent $i$,
\item $pos_i, rot_i \in \mathbb{R}^3$ represent agent $i$'s position and rotation, 
\item $e_i \in E$, $a_i \in A$, and $succ_i \in \{ 0, 1 \}$ respectively represent the object held by agent $i$, the action undertaken by agent $i$ in the last step, and whether the last action was a success or a failure.
\end{itemize}

\noindent \textbf{World.}
The world is denoted as $W = (S, A, T)$, wherein:
\begin{itemize}
\item $S = \{e_i, h_j\}_{i=1,..., N_o, j=1, 2}$ represents the world state, which encompasses the state of $N_o$ objects and $2$ agents;
\item $A = \{a_i\}$ denotes the action space;
\item $T = P(S'|S, A; \eta_i)$ refers to the transition matrix controlled by agent capability.
\end{itemize}

\noindent \textbf{Task.}
Given our formulation of agent capabilities, we select three everyday tasks for practical implementations of our challenge: \textit{Make Breakfast}, \textit{Arrange Room}, and \textit{Make Coffee}. These tasks respectively involves different capabilities: $(\alpha_i, \beta_i, \gamma_i, \delta_i, \epsilon_i, \zeta_i)$ for \textit{Cook Potato}, $(\alpha_i, \beta_i, \gamma_i)$ for \textit{Arrange Room}, and $(\alpha_i, \beta_i, \epsilon_i, \zeta_i)$ for \textit{Make Coffee}.

\noindent \textbf{Goal.}
Upon setting a task, the agent need to identify several key points (goals) that they must achieve. These goals are hidden from the helper, but affect the helper's reward for a particular action. Each goal is defined by the target state of the involved objects. The goal at time $t$ is represented as: 
\begin{equation}
\vspace{-0.1cm}
g^t = (\text{target}\_\text{state}^t, e^t, pr^t)
\label{eq:goal}
\end{equation}
Target state includes \textit{Wait}, \textit{Get}, \textit{On}, \textit{In}, \textit{KeepOpen}, \textit{KeepClose}, \textit{KeepOn} and \textit{KeepOff}, e.g., ``(In, Cup, Cabinet)''. 

\noindent \textbf{Observation Space.}
The proposed environment supports both symbolic and visual observations. This dual-mode observation system enables the helper to acquire and refine helping behaviors in various contexts. Following previous work~\cite{puig2020watch,puig2023nopa}, we use symbolic observations containing physical states of all perceivable objects and agents. Hidden states, like goals and capabilities, can not be observed. 

\section{Our Model}

Our objective is to train a helper agent who can effectively adapt to main agents with varying capabilities and goals. There are two main challenges: 1) the challenge posed by the environment (e.g., partial observation and occlusion), and 2) the challenge posed by the task (i.e., how to estimate the current goal $g^t$ and capabilities $\eta$ of the main agent). Given the observation $o^{0: t}$, the action policy for the helper is formulated as:

\vspace{-0.1cm}
\begin{equation}
  P(a|o^{0: t}) = \sum_{g^t \in G, \eta \in C} P(a|g^t, \eta, o^{0: t}) P(g^t, \eta|o^{0: t}).
  \label{eq:action_policy}
\end{equation}

Therefore, as depicted in ~\cref{fig:model}, our model consists of: 1) an opponent modeling module to estimate the current goal and capabilities of the main agent, i.e., $P(g^t, \eta|o^{0: t})$; and 2) a helping policy module to learn the distribution of the helper's action conditioned on the predicted goal and capabilities of the main agent, i.e., $P(a|g^t, \eta, o^{0: t})$. 

The training of our model has two stages. In the first stage, we train an opponent modeling module using the collected main agent trajectories. This stage is crucial as it enables the helper agent to understand and anticipate the behavior of the main agent effectively. In the second stage, we utilize the pre-trained opponent modeling module to train the helping policy module, which handles dynamic and complex tasks by interacting in our environment. 

\subsection{The Opponent Modeling Module}
\label{subsec:opponent_modeling_module}
To train this module, we collect simulated trajectories in our environment with the main agent following an expert policy and the helper moving randomly, where each frame contains the observations of the helper agent and the true labels of the main agent's goals and capabilities. Throughout data collection, we utilize a main agent with various tasks and capabilities and a helper executing random actions in the environment. See more details in the supplementary.

We use a sliding window trick when inferring goals and capabilities from the main agent's trajectory since goals are transient and typically persist for only a few steps. This method enables examining a segment of the trajectory at any given time, and sliding the window as new actions are taken. It allows the helper agent to make more immediate and relevant inferences about the main agent's current goals and capabilities, enhancing its adaptability and responsiveness. Formally, we have:

\begin{equation}
  P(\eta_i, g^t | o^{0:t}) \approx P(\eta_i, g^t | o^{t-w:t}),
  \label{eq:opponen_modeling}
\end{equation}
where $w$ is the window size. 

We use an object encoder and an agent encoder to extract features from raw observations (details in supplementary). Similar to the ``class token'' in ~\cite{alexey2020vit}, we prepend a learnable embedding, which we call ``object token'' to the sequence of object feature. We follow the usage of ``class token''~\cite{alexey2020vit}, and only retain this object token to represent the object feature $\xi^t_o$ after the transformer~\cite{vaswani2017attention} layer (other tokens are removed). This object feature $\xi^t_o$, together with the agent feature $\xi^t_a$, constitute the state feature $\xi^t_s$. As in ~\cref{eq:opponen_modeling}, we use a window size of $w$ and the state features $\{\xi^{t-w}_s, \cdots, \xi^t_s\}$ are fed into an MLP layer to obtain the final state feature $\xi_s$. Furthermore, a feature extractor is employed to derive the capability feature $\xi_{\eta}$ and the target state feature $\xi^t_g$ of the main agent. A transformer layer is utilized for the capability feature. As for the goal feature $\xi^t_g$, as detailed in ~\cref{eq:goal}, one transformer layer is used for the target state feature, and two MLP layers are employed for the features of the related object and its target parent receptacle.

\subsection{The Helping Policy Module}

Our helping policy module has an actor-critic~\cite{mnih2016actorcritic} architecture. At each time step $t$, given the current observation $o^{t}$, together with all the features $\xi_\eta$ and $\xi_g^t$ from the opponent modeling module, the helper agent needs to learn the helping policy $\pi_{\theta}(a|o^{t},\eta,g^t)$. We employ an MLP layer for action selection (the ``Policy MLP'' in \cref{fig:model}), coupled with an LSTM network for context memorization. Meanwhile, another MLP (the ``Value MLP'' in \cref{fig:model}) is utilized for the value network to estimate the value of the current state. 

\section{Experimental Setup}
\subsection{Two-stage Learning}
\label{subsec:help_policy_learning}

We apply a two-stage training technique to train our model. In the first stage, we use four auxiliary classifiers to respectively train the capability feature $\xi_\eta$, target state feature, object feature and parent receptacle feature (i.e., the goal feature $\xi^t_g$) in a supervised learning manner. These classifiers are lightweight and efficient, utilizing MLPs, and are trained using cross-entropy loss and the Adam optimizer, with a learning rate of $1\times 10^{-6}$ and a batch size of 32. 

In the second stage, we remove these four classifiers, keep the pre-trained opponent modeling module frozen, directly use the learnt opponent features $\xi_\eta$ and $\xi^t_g$, and focus on the training of helping policy module. We use 20 kitchen rooms for helping policy training and 10 for evaluation in AI2-THOR~\cite{kolve2017ai2}. The RLlib~\cite{liang2018rllib} framework and PPO~\cite{schulman2017ppo} algorithm are applied to train our helping policy module. Unless otherwise stated, we use the default settings of RLlib. During training, the scene is randomly initialized with a new room, a new task, and a new capability distribution of the main agent, at each episode. 

We apply a progressive learning technique to train the help policy module. Firstly, we train the helper policy module for 840 epochs, with a constant learning rate of $5\times 10^{-5}$ without weight decay and a batch size of 128, using the reward defined in ~\cref{eq:reward_new} with $\lambda_e=0.0$. We find such ``warm-up'' quite important in helping the helper agent to familiarize itself with basic skills to complete goals and tasks. Secondly, we progressively train the model for 240 epochs with a learning rate of $5\times 10^{-7}$ and $\lambda_e=1.0$ in~\cref{eq:reward_new} to enhance the smart help ability. In the evaluation phase, we test the helper in 10 rooms (never used in training) and with 14 different task-capability pairs, repeating three times with different random seeds. We exclude 7 task-capability pairs as they will enable the main agent to finish the task independently. See supplementary for more details.


\subsection{Reward}

The reward of the helper is influenced by the goal $g_m$ and the capability $\eta_m$ of the main agent. During training, the helper agent will get a reward of 20 after finishing a goal of the main agent, corresponding to the first term in ~\cref{eq:reward_new}. The second term $R^{e_m}(s, a_h)$ in \cref{eq:reward_new} is influenced by the capability of the main agent. Specifically, if the helper finishes a goal that the main agent can handle independently, the helper will get a punishment: $R^{e_m}(s, a_h)=-30$. For every step, the helper agent will get a cost of -0.12 normally and a cost of -0.5 if the action is illegal (e.g., attempting to open something that can not be opened, such as an apple) in the environment. If the task is not finished when the scenario is ended, the helper will receive a punishment of -20.


\subsection{Baselines} 

To make a fair comparison, the baselines are:
\begin{itemize}
    \item \textbf{Random}. This helper randomly selects an action. 
    
    \item \textbf{End2end-$\lambda_e$=$0.0$.} This model generates actions directly from the observation with MLPs, trained with $\lambda_e=0.0$ in ~\cref{eq:reward_new}. We set learning rate to $5\times 10^{-5}$, batch size to 128. The model is trained for 350 epochs before convergence. 
    
    \item \textbf{End2end-$\lambda_e$=$1.0$.} Similar to \textit{End2end-$\lambda_e$=$0.0$}, but with $\lambda_e=1.0$ in ~\cref{eq:reward_new}. 

    \item \textbf{MCTS.} This model uses a Monte-Carlo Tree Search (MCTS~\cite{browne@mcts}) algorithm to search for an action with the highest value, given a predicted goal from the opponent modeling module. The value of each action is estimated by the status of goal completion after rollout, discounted by the number of steps taken to achieve this result.
    
    \item \textbf{MCTS-heuristic.} This model combines the \textit{MCTS} model with expert rules for the three selected tasks. Given a predicted goal from the opponent modeling module, it decides on which action to simulate at a probability $p=0.5$ by chance and $p=0.5$ by a rule-based selection of the next action to complete the goal.
    
    \item \textbf{MCTS$_{\text{TG}}$.} This is an implementation to reproduce the high-level planning policy of Watch-and-Help~\cite{puig2020watch}. This model knows the true goal of the main agent. 
    
    \item \textbf{MCTS$_{\text{RG}}$.} This model is almost the same as \textit{MCTS$_{\text{TG}}$}, except that it follows a random goal.
    
    \item \textbf{LLM.} We use Large Language Model, i.e., \textit{gpt-3.5-turbo-instruct}~\cite{openai@gpt3x5} as the helper. We compose a prompt based on the state observations from the AI2-THOR~\cite{kolve2017ai2}. The prompt is fed into the LLM at every step. We extract the action decision generated from the LLM and evaluate its actual effectiveness in the AI2-THOR simulator. See supplementary for more details.
\end{itemize}

\subsection{Ablation Study} 

To assess the contributions and efficacy of essential components in our method, we derive the following variants:
\begin{itemize}
    \item \textbf{BaseModel.} This model is the aforementioned helping policy model with a pre-trained parameter-frozen opponent modeling module. As in~\cref{subsec:help_policy_learning}, we set $\lambda_e=0.0$ in~\cref{eq:reward_new}, and train it for 840 epochs.

    \item \textbf{BaseModel-w/o. Capability.} Here we remove the capability embedding from the training of the helping policy.
        
    \item \textbf{BaseModel-$\lambda_e$=$1.0$.} This model is trained with $\lambda_e=1.0$ in ~\cref{eq:reward_new} for 840 epochs.

    \item \textbf{BaseModel-PL-$\lambda_e$=$0.0$.} Here, ``PL" means the Progressive Learning technique (detailed in \cref{subsec:help_policy_learning}), but with $\lambda_e=0.0$ in the second phase after the first warm-up phase. We simply continue to train the \textit{BaseModel} with a smaller learning rate for 240 epochs (with no change in $\lambda_e=0.0$).
    
     \item \textbf{BaseModel-PL-$\lambda_e$=$1.0$ (our full model).} We use the correct Progressive Learning technique (detailed in \cref{subsec:help_policy_learning}), and change to $\lambda_e=1.0$ in the second phase after warming up the basic skills of the helper model.
     
\end{itemize}

\begin{figure*}[t!]
\begin{center}
   \includegraphics[width=1.0\textwidth]{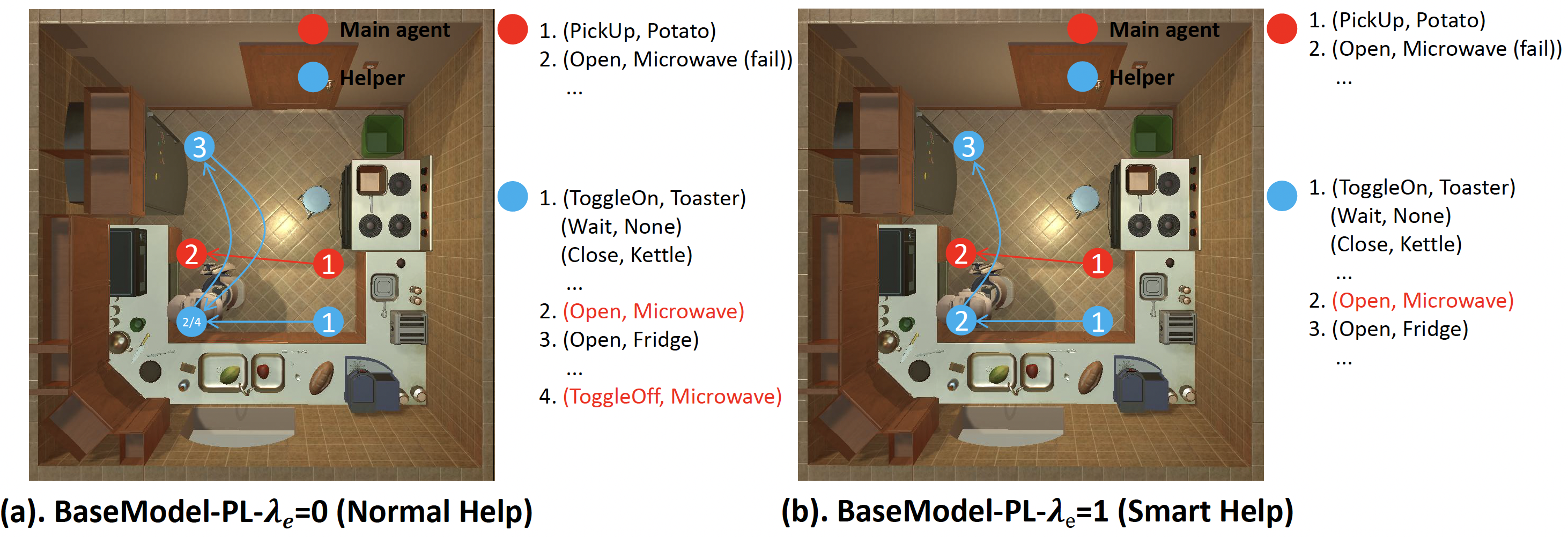}
\end{center}
\vspace{-0.4cm}
   \caption{Qualitative results of the learned \textit{smart help} policy. In this example, the main agent moves directly to finish the goal ``(In, Potato, Microwave)'', but fails at the goal ``(Open, Microwave)''. (a) The \textit{BaseModel-PL-$\lambda_e$=$0.0$} model not only helps the main agent with the bottleneck ``(Open, Microwave)'', but also continues to help with another goal ``(ToggleOff, Microwave)''. (b) Our full model \textit{BaseModel-PL-$\lambda_e$=$1.0$} only offer necessary help with the bottleneck ``(Open, Microwave)'', and let the main agent to finish the rest goals independently, which helping policy is smarter since it considers the needs and emotional feelings of the main agent.  
    }
\label{fig:qua_res}
\vspace{-0.4cm}
\end{figure*}

\subsection{Evaluation Metrics} 

To objectively evaluate an agent's performance, we utilize six distinct metrics, where \textit{Helping Necessity (HN)} and \textit{Helping Rate (HR)} are first proposed in our work to better assess the \textit{Smart Help} policy. Let $N$ denote the size of the test scenarios, and let each scenario $i$ be characterized by an initial room state $s^0_i$ and a goal of transitioning the room to the target state $s^*_i$. Assuming that both the main agent and the helper require $l_i$ steps (with a maximum of 30) to reach the final state $s^l_i$, we elaborate on these metrics as follows.

\begin{itemize}
\item  \textbf{Success Rate (SR and GSR).} It includes \textit{Task-conditioned Success Rate (SR)} and \textit{Goal-conditioned Success Rate (GSR)}, respectively representing the completion degree of tasks and goals. The \textit{SR} is defined as $SR=\frac{1}{N}\sum_{i=1}^N R_i$, where $R_i$ is 1 if the task is finished. For task $i$, let $N_{g^{m}_i}$ denote the effective goals completed by the main agent, $N_{g^{h}_i}$ represent the effective goals completed by the helper, and $N_{g^{a}_i}$ be all the effective goals of the task. The \textit{goal-conditioned success} for task $i$ is: $GS_i = \frac{N_{g^{m}_i}+N_{g^{h}_i}}{N_{g^{a}_i}}$. Hence, the overall \textit{GSR} is defined as $GSR = \frac{1}{N}\sum_{i=1}^N GS_i$. $R_i = 1$ if and only if $GS_i = 1$. 

\item \textbf{Helping Necessity (HN).} This metric reflects the necessity of the helper's involvement. 
Only when the capability required to finish the goal exceeds the main agent's capability, the helping is necessary. 
Let $N_{g^{h}_{\text{nec}, i}}$ denote the necessary goals completed by the helper, the Helping Necessity is then calculated as $HN = \frac{1}{N}\sum_{i=1}^N \frac{N_{g^{h}_{\text{nec}, i}}}{N_{g^{h}_i}} R_i$, where $R_i=1$ if the task is completed, otherwise $R_i=0$. This assesses the helper's ability to swiftly identify the main agent's capability and provide necessary assistance. 
\item \textbf{Helping Rate (HR).} \textit{HR} represents the helper's initiative to help. $HR = \frac{N_{\text{help}}}{N_{\text{need}\_\text{help}}}$, reflecting the probability of helping when the main agent needs help. 
\item \textbf{Episode Length (EL).} We record the average \textit{episode length} to reflect the efficiency of helping policy. It is computed as: $EL = \frac{1}{N}\sum_{i=1}^{N}l_i$.
\item \textbf{Success-weighted by Path Length (SPL).} We also include \textit{SPL} to have a comprehensive evaluation. It is computed as: $SPL = \frac{1}{N}\sum_{i=1}^{N}R_i\frac{d_i}{\max(d_i, l_i)}$, where $R_i\in \{0,1\}$ denotes whether the task is successfully completed, $d_i$ represents the minimum number of steps to finish the task $i$, and $l_i$ is the actual steps.  
\end{itemize}

Beyond these key metrics, we also incorporate the \textit{average rewards} as a metric for evaluation. For the \textit{Smart Help} challenge, we want to increase the \textit{HN} and \textit{HR}, while simultaneously descending the \textit{EL}, as well as keeping the \textit{SR}, \textit{GSR}, and \textit{SPL} as high as possible. 

\section{Results and Analyses}

\noindent \textbf{Comparison with baselines.}
As shown in ~\cref{tab:performance}, our full model exhibits superior performance compared with all other baseline models without ground truth knowledge of the main agent. 
The \textit{Random} agent, selecting random actions, and the \textit{MCTS$_{\text{RG}}$} model, following random goals, both have minimal ability to provide appropriate help. 
The superiority of our \textit{BaseModel} compared to the \textit{End2End} model underscores the value of opponent modeling in the context of assistance tasks. 
The \textit{LLM} agent, although has the shortest \textit{EL}, falls short on other metrics compared with our model. 
The \textit{MCTS-heuristic} model, which incorporates rule-based heuristics, outperforms the pure \textit{MCTS} model. However, our model achieves higher scores in \textit{SR}, \textit{GSR}, \textit{HN}, and \textit{HR}, while maintaining similar values for \textit{EL} and \textit{SPL}, indicating the effectiveness of its assistive action planning. The \textit{MCTS$_{\text{TG}}$} model, knowing the true goal of the main agent, achieves the best performance among all the models, serving as an upper bound for the task. Notably, while it surpasses other models in most metrics, our model demonstrates competitive performance, particularly in $HN$. This suggests that our model achieves a balance by providing only necessary assistance and ensuring user comfort. The gap between our model and the upper bound shows potential for future research and further improvement.

\noindent \textbf{Ablation analysis.} Comparing the \textit{BaseModel} with \textit{BaseModel-$\lambda_e$=$1.0$}, we find that setting $\lambda_e=0.0$ can augment the assisting actions of the helper. When compared with \textit{BaseModel-w/o. Capability}, we find that the capability module contributes to the learning of the smart helping policy. Comparing the \textit{BaseModel-PL-$\lambda_e$=$0.0$} with \textit{BaseModel-PL-$\lambda_e$=$1.0$}, we find that after the initial ``warm-up'' of the basic skills, setting $\lambda_e=1.0$ in the second phase could greatly improve the \textit{HN} while maintaining competitive performances across other metrics. Thus, as in \cref{subsec:help_policy_learning}, for our full model, in the \textit{BaseModel} ``warm-up'' training phase, the model learns how to complete the goals and tasks, and the assistive actions are greatly encouraged; while in the second progressive learning phase, our full model focuses on improving the assistance strategy of the helper based on the main agent's needs and feelings.

\noindent \textbf{Qualitative Results.} 
\cref{fig:qua_res} demonstrates two helping cases with a baseline model and our full model. The two models exhibit exploration behaviors in the scene and finally learn to solve the bottleneck problem of the main agent. In comparison, the \textit{BaseModel-PL-$\lambda_e$=$0.0$} only learns to help all the goals it successfully infers, while our full model \textit{BaseModel-PL-$\lambda_e$=$1.0$} learns to decide whether to help based on its inferred goals and capabilities, i.e., only to help with ``(Open, Microwave)''.

\begin{table}[t!]
    \centering
    \vspace{-0.05cm}
    \resizebox{\linewidth}{!}{%
        \begin{tabular}{l c c c c c c c}
            \toprule
    \rowcolor{mygray} Method & SR($\uparrow$) & GSR($\uparrow$) & HN($\uparrow$) & HR($\uparrow$) & Reward($\uparrow$) & EL($\downarrow$) & SPL($\uparrow$)\\ 
   \midrule
   Random& 0.074 & 0.364 & 0.054 & 0.057 & -31.339 & 28.567 & 0.054 \\
   End2end-$\lambda_e$=$0.0$& 0.267 & 0.441 & 0.149 & 0.315 & -23.218 & 24.343 & 0.212 \\
   End2end-$\lambda_e$=$1.0$ & 0.067 & 0.372 & 0.057 & 0.057 & -26.183 & 28.726 & 0.049 \\
   LLM & 0.345 & 0.506 & 0.382 & 0.381 & / & \textbf{22.914} & 0.210 \\
   MCTS& 0.178 &0.482 &0.266 &0.328 &/&26.743 	&0.126 
    \\
   MCTS-heuristic& 0.274 &0.528 &0.355 &0.400 &/&24.692 	&0.199    \\
   $\text{MCTS}_{\text{TG}}$ &  \underline{0.593} & \underline{0.804} 	& \underline{0.610} & \underline{0.678} &/& \underline{16.271} & \underline{0.541} 
 \\
   $\text{MCTS}_{\text{RG}}$ &0.043 &0.348 	&0.015 &0.015 & / &28.950 &0.042  \\
   \hline
   BaseModel& 0.419 & 0.515 & 0.408 & 0.412 & -15.179 & 24.945 & 0.176 \\
   BaseModel-w/o. Capability& 0.374 & 0.504 & 0.375 & 0.379 & -13.852 & 23.748 & \textbf{0.235} \\
   BaseModel-$\lambda_e$=$1.0$& 0.181 & 0.466 & 0.193 & 0.200 & -17.185 & 27.448 & 0.122 \\
   BaseModel-PL-$\lambda_e$=$0.0$& \textbf{0.488} & \textbf{0.556} & 0.455 & 0.496 & -10.959 & 24.383 & 0.198 \\
   \cellcolor{mygray}\textbf{BaseModel-PL-$\lambda_e$=$1.0$ (Ours)}& \cellcolor{mygray}0.483 & \cellcolor{mygray}0.548 & \cellcolor{mygray}\textbf{0.498} & \cellcolor{mygray}\textbf{0.506} & \cellcolor{mygray}\textbf{-10.625} & \cellcolor{mygray}24.650 & \cellcolor{mygray}0.200 \\
    \bottomrule
    \end{tabular}%
    }
    \vspace{-0.25cm}
    \caption{
        The quantitative results of our experiments. The last ten environments in AI2-THOR are reserved for evaluation. The \textbf{best results} are highlighted in bold. Note that we include some oracle baselines and the \underline{upper bound performances} are highlighted with underlines. 
    }
    \vspace{-0.55cm}
    \label{tab:performance}
\end{table}

\section{Conclusion}

In this paper, we propose the novel challenge \textit{Smart Help} and build an environment and model that fosters smarter and more harmonious interaction between humans and artificial agents. We hope our challenge, environment, dataset, model and benchmark results will serve as valuable resources to future studies of this important problem.



\paragraph{Acknowledgement} This work is supported by the National Science and Technology Major Project (2022ZD0114900). We thank Zhenliang Zhang at BIGAI and Luyao Yuan at META for their helpful discussions. 

{
    \small
    \bibliographystyle{ieeenat_fullname}
    \bibliography{main}
}


\clearpage
\setcounter{page}{1}
\setcounter{figure}{0}
\setcounter{table}{0}

\maketitlesupplementary
\appendix

\section*{Overview}

This supplementary material includes: 
\begin{itemize}
\item \cref{sec:environment} has the detailed setting of the environment.
\item \cref{sec:implementation} has the model implementation details.
\end{itemize}

\section{Environment Setting}
\label{sec:environment}


\begin{figure*}[t]
\begin{center}
   \includegraphics[width=0.87\textwidth]{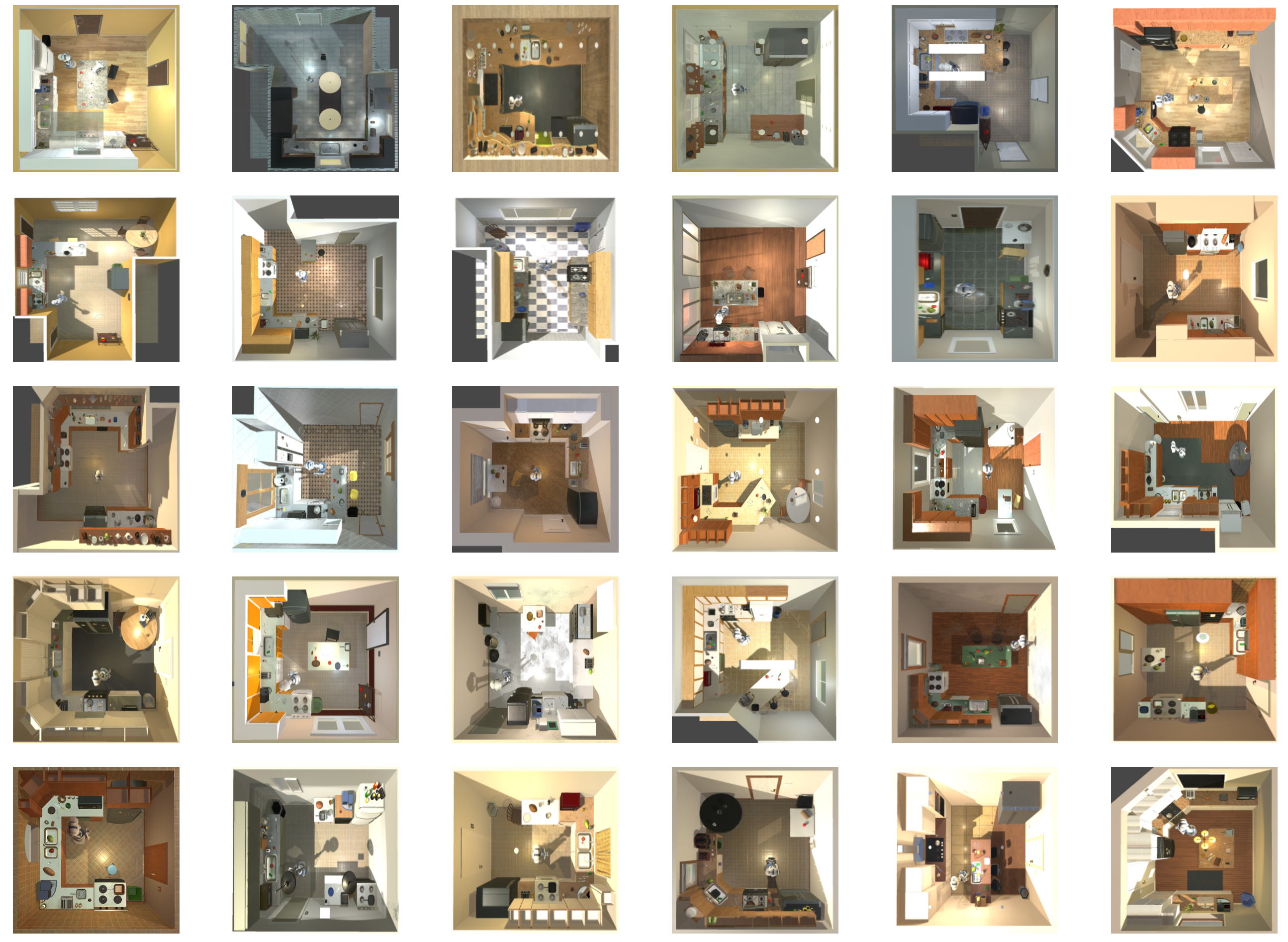}
\end{center}
\vspace{-0.4cm}
   \caption{
   In our study, a total of 30 distinct scenes from AI2-THOR were utilized to train our model. 
   The planforms of them are provided here. 
   These scenes were divided into two datasets: 20 for training while the remaining 10 were reserved for testing. 
   Different scenes have different size, but the figures presented here are scaled to the same size. 
   The object composition in each scene is diverse. Thus the agent must learn a general policy for different scenes. 
   However, for our tasks, the focus was solely on those objects that were common across all scenes. This approach was adopted to ensure that the task was consistently applicable across all the scenes used in the study. }
\label{fig:scene}
\vspace{-0.4cm}
\end{figure*}

\paragraph{Scenes}
We use 30 scenes from AI2THOR kitchen scenes, which are shown in \cref{fig:scene}. In these scenes, 20 are used for training while the remaining 10 are reserved for testing. The object composition in each scene is diverse. We only pay attention to objects that were common across all scenes when parsing the task to goals. 

\paragraph{Tasks} 
We have classified our tasks into three categories: MakeBreakfast, ArrangeRoom, and MakeCoffee. Each task is further divided into a series of goals. The  \cref{tab:parse} shows some examples about the parsing process. Different room will leads to different parsing. For example, in some rooms the mug has been already put in the coffee machine, thus the main agent does not need to pick up the mug and put it in the coffee machine. 

\begin{table}
    \centering
    \resizebox{\linewidth}{!}{%
        \begin{tabular}{l c}
            \toprule
    Task& Goals (example)\\ 
   \midrule
   MakeBreakfast& (Get, Potato), (KeepOff, Mircowave), \\
   & (KeepOpen, Mircowave), (In, Potato, Microwave), \\
   & (KeepClose, Microwave), (KeepOn, Microwave), \\
   &  (KeepOff, Microwave)\\
   \hline
   ArrangeRoom&  (Get, Potato), (KeepOpen, Fridge), \\
   & (In, Potato, Fridge), (KeepClose Fridge), \\
   \hline
   MakeCoffee& (Get, Mug), (In, Mug, CoffeeMachine), \\
   & (KeepOn, CoffeeMachine), (KeepOff, CoffeeMachine)\\
   \bottomrule
    \end{tabular}%
    }
    \caption{
    The table showcases examples of the parsing process, from overarching tasks to specific goals. This process may vary across different scenes and may also be probabilistic within the same scene. For instance, in certain scenarios, the mug may already be situated in the coffee machine. In such cases, the agent only needs to activate and then deactivate the machine to complete the ``MakeCoffee'' task.}
    \label{tab:parse}
\end{table}

\paragraph{Goals}
Goals denote the target states that an agent must reach before successfully completing a task. We have identified seven potential goals for the agent to choose from. These are: (Wait), (Get, $e_i$), (On, $e_i$, $pr_i$), (In, $e_i$, $pr_i$), (KeepOpen, $e_i$), (KeepClose, $e_i$), (KeepOn, $e_i$), and (KeepOff, $e_i$). 
In certain scenarios, the sequence of goals is pre-determined. This means that the agent must accomplish the goals in the given order, rather than arbitrarily. For instance, the microwave must be closed before it can be toggled on. Failing to follow this sequence results in an invalid action, as a microwave cannot be activated while open.
Moreover, once a specific state has been reached, certain properties of that state need to be maintained until the next goal indicates a change. For example, if the goal sequence comprises (Open, Fridge) followed by (Put, Bread, Fridge), the agent must keep the fridge open after the first goal is achieved, until successfully accomplishing the ensuing goal of placing the bread in the fridge.
These complexities present challenges both in the parsing process, which converts tasks into goals, and in the execution of these goals, adding layers of difficulty to the agent's tasks.

\paragraph{Intentional action}

We have defined seven intentional actions that the agent can perform: (Wait), (PickUp, $e_i$), (Put, $pr_i$), (ToggleOn, $e_i$), (ToggleOff, $e_i$), (Open, $e_i$), and (Close, $e_i$). Upon selecting an intentional action, a low-level planner devises a sequence of executable actions to be sent to the AI2THOR simulator.
It's important to note that actions may fail due to various reasons. For instance, the action (Put, $pr_i$) will result in failure if the agent is not currently holding any objects. Similarly, an action to open an item that cannot be opened is destined to fail.
Furthermore, the validity of an action can be context-specific. For example, while a microwave can typically be opened, it cannot be opened if it is currently active; it must first be toggled off. This context-dependent validity of actions presents a significant challenge for the model. It necessitates the model to learn the intricate dependency relationships between the context and the action, based on the reward and observation data provided by the environment. 

\begin{figure*}[t]
\begin{center}
   \includegraphics[width=0.87\textwidth]{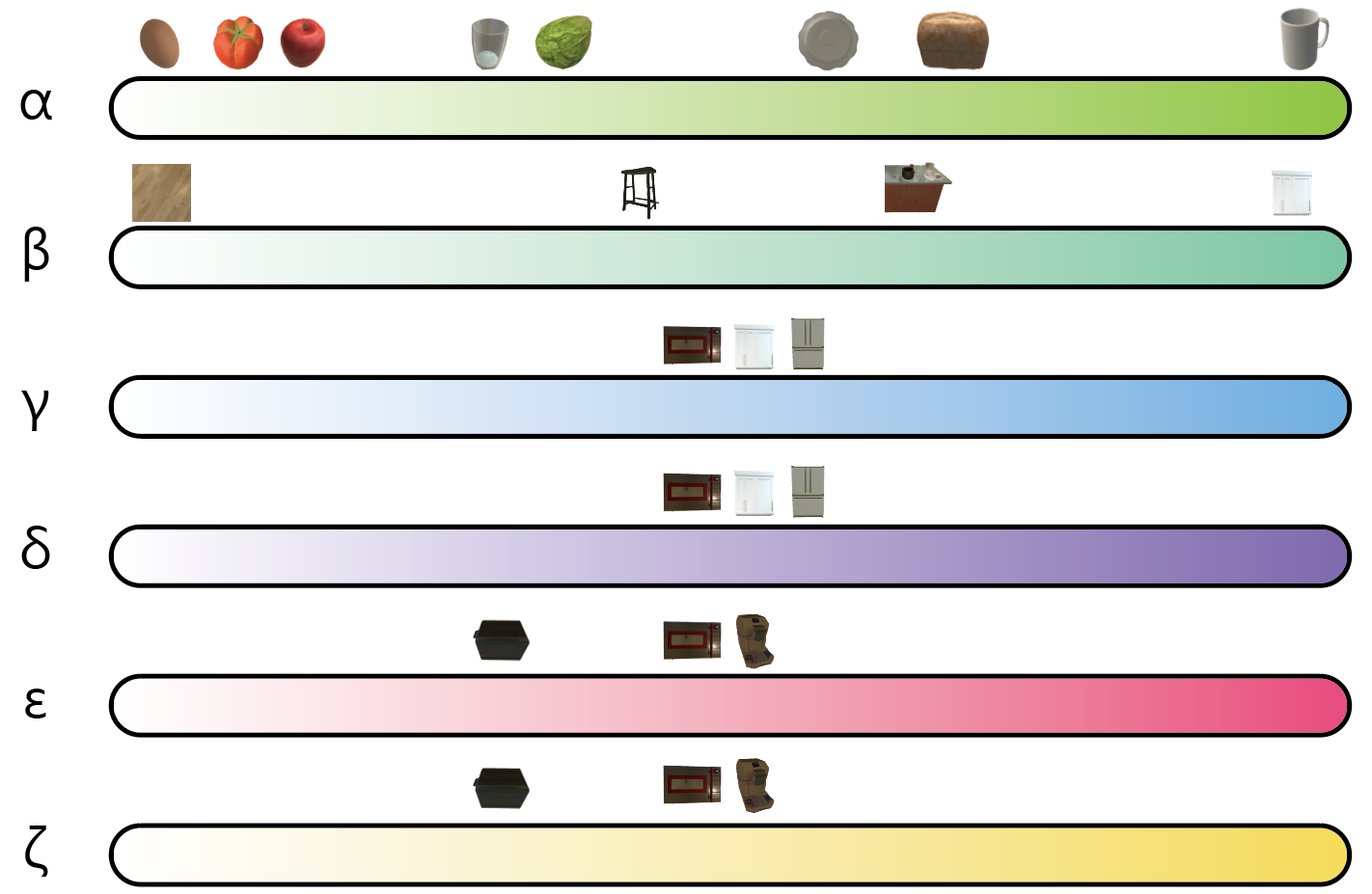}
\end{center}
\vspace{-0.4cm}
   \caption{
    This figure shows the impact of different capabilities, represented by $\alpha$, $\beta$, $\gamma$, $\delta$, $\epsilon$, and $\zeta$, on the agent's interaction with various objects. 
    $\alpha$ denotes the agent's ability to lift objects of different weights. A lower $\alpha$ value implies a limited capacity to handle only lighter objects, such as a potato (weight 0.18) or a tomato (weight 0.20). However, as $\alpha$ increases, the agent can manage heavier items, such as bread (weight 0.7) or a cup (weight 1.0). These object weights are determined by the AI2THOR simulator \citep{kolve2017ai2}, reflecting the careful considerations of its developers. 
    $\beta$ symbolizes the agent's ability to reach varying heights. The floor, with a height of 0, is the most accessible, while the other objects' height, like the cabinet's height, determined by its presence in the scene, poses a greater challenge. 
    $\gamma, \delta, \epsilon$ and $\zeta$ respectively represent the agent's ability to perform specific operations: opening, closing, toggling on, and toggling off. We assume a threshold of 0.5, which serves as the tipping point between success and failure in these actions under normal circumstances. }
\label{fig:capability}
\vspace{-0.4cm}
\end{figure*}

\paragraph{Capability}
Some examples about how the capability will affect agent's interaction with different objects are shown in \cref{fig:capability}. 
We represent the capability by $\alpha$, $\beta$, $\gamma$, $\delta$, $\epsilon$, and $\zeta$. 
\begin{itemize}
\item $\alpha_i \in [0, 1]$ denotes the maximum height that the agent can reach;
\item $\beta_i \in [0, 1]$ represents the maximum weight that the agent can lift;
\item $\gamma_i, \delta_i, \epsilon_i, \zeta_i \in [0, 1]$ represent the agent's ability to complete open, close, toggle on, and toggle off tasks, respectively.
\end{itemize}

In this challenge, we have seven types of agents to select from. These include agents with full capabilities and agents with one randomly assigned limitation. An agent with full capabilities will have $(\alpha, \beta, \gamma, \delta, \epsilon, \zeta) = (1, 1, 1, 1, 1, 1)$.
If the agent has a limitation in $\alpha$, the value will fall within the range of $(0.2, 0.8)$. The same concept applies to other capabilities, where $\beta$ will be within $(0.1, 0.7)$, and $ \gamma, \delta, \epsilon, \zeta$ will range from $(0, 0.49)$.

In the training process, the main agent will be randomly set to a kind of capability. In the testing process, we will set the main agent capability to the lowest level. For example, when the main agent's type is $\beta$ limitation, his capability is (1, 0.1, 1, 1, 1, 1). 
  
\begin{table}[t!]
    \centering
    \resizebox{\linewidth}{!}{%
        \begin{tabular}{l c}
            \toprule
    Type & Capability\\ 
   \midrule
   Full capability & (1, 1, 1, 1, 1, 1) \\
   $\alpha$ limitation & ($\alpha$, 1, 1, 1, 1, 1), $\alpha \in (0.2, 0.8)$ \\
   $\beta$ limitation & (1, $\beta$, 1, 1, 1, 1), $\beta \in (0.1, 0.7)$ \\
   $\gamma$ limitation &  (1, 1, $\gamma$, 1, 1, 1), $\gamma \in (0, 0.49)$\\
   $\delta$ limitation &  (1, 1, 1, $\delta$, 1, 1), $\delta \in (0, 0.49)$ \\
   $\epsilon$ limitation & (1, 1, 1, 1, $\epsilon$, 1), $\epsilon \in (0, 0.49)$ \\
   $\zeta$ limitation & (1, 1, 1, 1, 1, $\zeta$), $\zeta \in (0, 0.49)$ \\
   \bottomrule
    \end{tabular}%
    }
    \caption{
    We sample agents from seven different types. 
    The types also include the sampling of agents with full capabilities, serving as a benchmark to examine the model's ability to differentiate between agents with limitations and those without. The range of capabilities for an agent with limitations is established based on the statistical information derived from the scenes. }
    \label{tab:capability}
\end{table}

\section{Implementation Details}
\label{sec:implementation}

\subsection{Trajectory dataset}

We assembled a main agent trajectory dataset for pretraining the opponent modeling module. The pretraining phase equips the model with a more effective starting point for the subsequent reinforcement learning (RL) training.
During data collection, we employ a human-derived expert policy to guide the main agent's actions. Simultaneously, the helper will randomly choose an action, with its observations and the ground label constituting the dataset. 
This dataset comprises 60,024 trajectories, each containing observations at five discrete time points and the final goal of the main agent. 
When at all of the five discrete time points the helper can not observe the main agent, the label of the main agent's goal will be set to None. 
The statistics of the dataset is shown in \cref{tab:data}. 

\begin{table}[t!]
    \centering
    \resizebox{\linewidth}{!}{%
        \begin{tabular}{l c c c c c c c c c c}
            \toprule
    Goal & KeepClose & Get & In & On & Open & ToggleOff & ToggleOn & Wait & None\\ 
   \midrule
   \# Trajectory & 4192 & 8295 & 3420 & 214 & 5066 & 2384 & 2337 & 12953 & 21154\\
   \bottomrule
    \end{tabular}%
    }
    \caption{
    \textbf{Dataset Statistics.}
    Our dataset is characterized by an imbalance in the number of trajectories corresponding to each goal. This discrepancy is addressed during the training process through a rebalancing strategy. The frequency of ``On'' goals is significantly lower compared to the others. This is primarily due to the interchangeable use of ``In'' goals to represent ``On'' goals. This substitution is reasonable as the agent can achieve them via the same singular ``Put'' action. This substitution is used particularly during the parsing of the task ``ArrangeRoom'', which aids in maintaining uniform terminology across the parsing.}
    \label{tab:data}
\end{table}

\subsection{Opponent Modeling Module}

We train an opponent modeling module utilizing the main agent trajectory dataset. 

First, we need to change the observations to embedding with the object encoder. 
We use MLPs and embedding model to achieve this. 
\begin{table}[]
    \centering
    \begin{tabular}{c|c|c|c}
    \hline
        Property & Data type & Method & Output size \\
        \hline
        Object type & Int & Embedding & 32 \\
        Parent receptacle & Int & Embedding & 32 \\
        Height & Float & MLP & 32 \\
        Weight & Float & MLP & 32 \\
        Position & Float & MLP & 32 \\
        Distance & Float & MLP & 32 \\
        Other properties & Bool & MLP & 32 \\
        \hline
    \end{tabular}
    \caption{The properties of object and how to change them to embedding. Here other properties include 'isPickedUp', 'isOpen', 'isCooked', 'isToggled', and 'isVisible'. }
    \label{tab:1}
\end{table}

After the change, we will get a feature will size 224 for every object. 
Then we use a transformer to handle all the features. 
The transformer has 8 heads, 0.2 dropout, hidden size of 128, and 4 layers. 
With the transformer, we get the object feature. 

For dealing with the observation of agents, we use the agent encoder \ref{tab:2}.  
\begin{table}[]
    \centering
    \begin{tabular}{c|c|c|c}
    \hline
        Property & Data type & Method & Output size \\
        \hline
        Held object type & Int & Embedding & 32 \\
        Action type & Int & Embedding & 32 \\
        Action success & Bool & MLP & 32 \\
        Position & Float & MLP & 32 \\
        Rotation & Float & MLP & 32 \\
        \hline
    \end{tabular}
    \caption{The observations of objects and how to change them to embedding. }
    \label{tab:2}
\end{table}
After the model processing, we concatenate the feature of action type and action success and put them into a MLP to get the action feature. 
Then we use other three features and the action feature to get the agent feature with MLP. 

The object feature and the agent feature are concatenate to form the state feature. 
Next, we concatenate state features of five time points and put them into a MLP. 
The time feature output by the MLP has size of 128. 
Then, the time feature is fed into two transformers, and then we will get two features, which represent goal and capability respectively. 
The transformer has two heads, 0.1 dropout, hidden size of 256, and 4 layers. 
The parameters of these two transformers are not shared. 
Then we concatenate the time feature and the goal feature and put them to a MLP. 
The we get the feature of $o_1$, whose size is 128. 
Then we concatenate the time feature, the goal feature and the $o_1$ feature and get $o_1$ feature with a MLP. 
The $o_1$ feature represents the object that the goal involves. 
The $o_2$ feature represents the parent receptacle that the goal involves. 

In the training process, we use four MLP as classifiers to get the probability of goals, the relevant objects, and the predicted capabiities. 
The we use cross-entropy loss and the Adam optimizer, with a learning rate of $1\times 10^{-6}$ and a batch size of 32. 

We achieved a commendable prediction accuracy of 83.6\%. This high level of precision indicates that our model possesses a remarkable capability to infer the goals of human agents based on a constrained set of observational data.

\subsection{Helping Model}

Our model accepts symbolic observations as input, which encompass the helper's partial observation, as well as the symbolic agent states of both the helper and the main agent. The observation includes the symbolic states of all objects within the helper's field of view.
The object encoder and the agent encoder have been introduced in the last part. 
With the assistance of our pre-trained opponent modeling module, we can effectively model the main agent using its trajectory.
It is worth noting that the classifiers from the opponent modeling module are not utilized within the helper model. Instead, the latent representation from the opponent modeling module is concatenated with the observation. This composite embedding is then processed through a linear layer and a ReLU function before being input into a single-layer LSTM with 512 hidden units.
To generate an action policy, we apply a linear layer to the output of the LSTM, with no additional nonlinearity required. 
This streamlined approach ensures the effective processing of symbolic observations and accurate policy generation.


\subsection{LLM}

\begin{figure*}[t]
    \centering
    \includegraphics[width=1.0\linewidth]{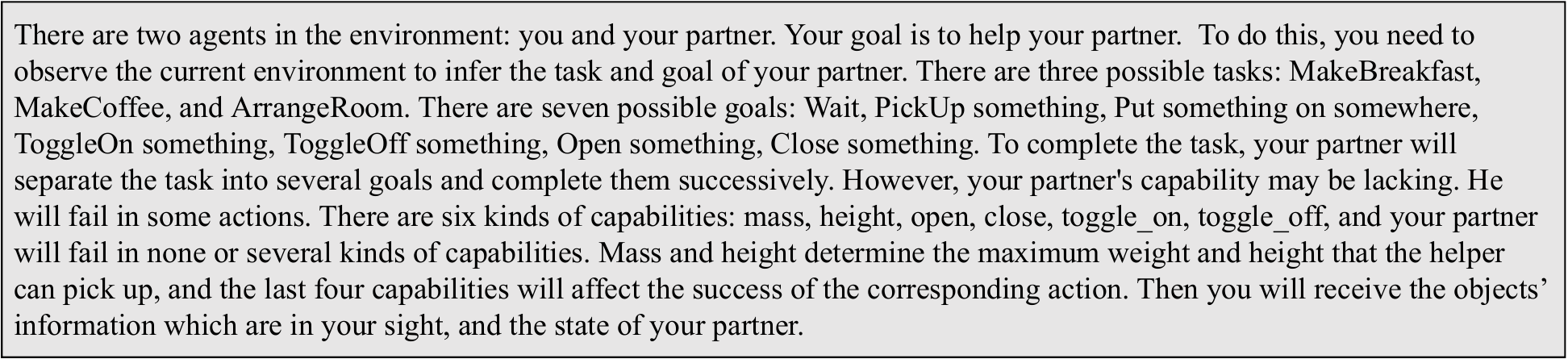}
    \caption{
        \textbf{The LLM prompt (the first part)}. We provide the basic setting information for the LLM, including what needs to be completed, what can be done, what types of the main agents are in the environment, etc.
    }
    \vspace{-0.3cm}
    \label{fig:prompt_llm_1}
\end{figure*}

\begin{figure*}[t]
    \centering
    \includegraphics[width=1.0\linewidth]{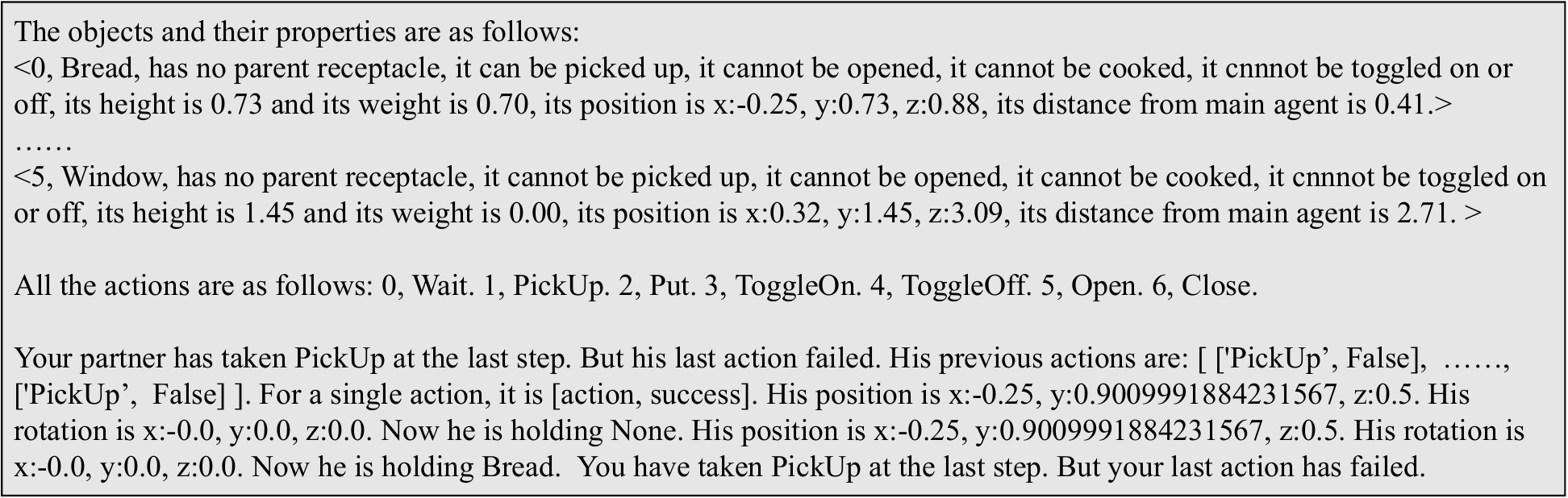}
    \caption{
        \textbf{The LLM prompt (the second part)}. At each step, we provide the LLM with its partial observations. For scene observations, we enumerate all observable objects and their properties. For agent observations, we summarize the action trajectory of the main agent, the completion of his previous action, and whether this main agent holds an object. All the optional actions and the previous action information of the LLM will also be mentioned.
    }
    \vspace{-0.3cm}
    \label{fig:prompt_llm_2}
\end{figure*}

\begin{figure*}[t!]
    \centering
    \includegraphics[width=1.0\linewidth]{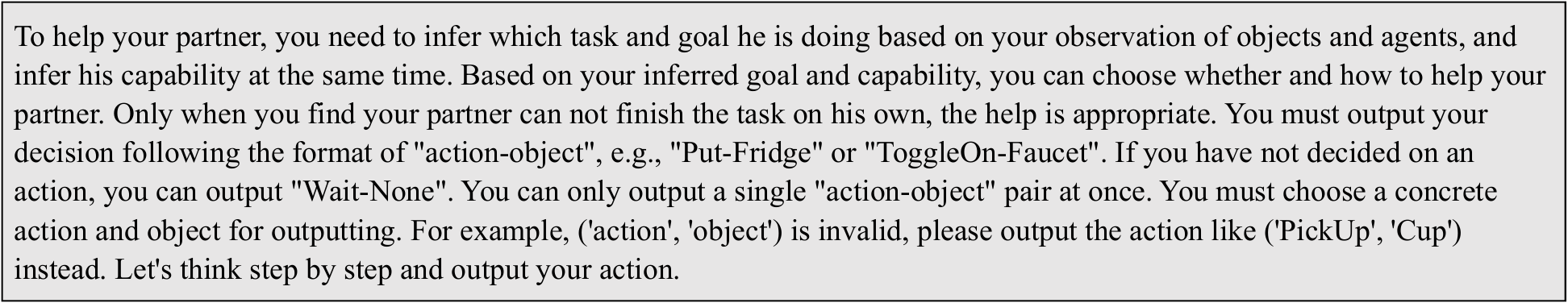}
    \caption{
        \textbf{The LLM prompt (the third part)}. We will specify the criteria for selecting a valid action through the LLM's output, along with providing illustrative examples to facilitate its comprehension.
    }
    \vspace{-0.3cm}
    \label{fig:prompt_llm_3}
\end{figure*}

In our setup, LLM, i.e., ``gpt-3.5-turbo-instruct'' is used as a planner to generate a policy for the helper agent. For each scene, we provide the basic information of this environment, as shown in \cref{fig:prompt_llm_1}. At each step, we give the partial observations of scenes and agents to the LLM, as shown in \cref{fig:prompt_llm_2}. Besides, as shown in \cref{fig:prompt_llm_3}, we stipulate how the LLM should output to select a valid action, and give an example to help it understand. The LLM processes the prompt and produces an action decision based on its understanding of the given information. If the LLM outputs an invalid action decision, the ``Wait'' action will be adopted as its next action. To assess the effectiveness of the LLM-generated actions, we execute this action in the AI2-THOR simulator to determine how well the LLM performs as a planner for the helper agent in the given environment. 

\subsection{MCTS-heuristic}
\begin{itemize}
    \item \textbf{MCTS.} We implement a MCTS algorithm which outputs an action based on a given goal, whether predicted, ground truth, or randomly selected. In each simulation, it samples an action under the guidance of the Upper Confidence Bound (UCB), which balances exploration and exploitation according to node values and visit counts. Then it performs a rollout to reach one possible end state, with the maximum rollout depth limited to 5 steps. If the goal is successfully achieved within this depth, the end state's value is evaluated by:
    $$v = 50/d$$
    where $v$ represents the value and $d$ the number of steps taken to complete the goal. The algorithm returns the action represented by the most visited node after $n=500$ simulations.

    \item \textbf{MCTS-heuristic.}To enhance search efficiency and task completion capabilities, we use a heuristic MCTS method, which incorporates hand-written rules that are specifically defined to break down a goal into a sequence of actions necessary for task completion. This model employs a probabilistic strategy during the sampling of the MCTS, with a probability parameter $p_{\text{sample}}$ determining whether to sample a random action for rollout or, at $1 - p_{\text{sample}}$, select the next action from the rule-based sequence, excluding completed actions. When $p_{\text{sample}}$ is set to $1$, this model is equivalent to the \textit{MCTS} model, and when it is set to $0$, it strictly follows the rule-based action sequence. Initially, we set $p_{\text{sample}}$ to $1$ and $0.9$ to encourage exploration for more efficient action sequences. The action sequences obtained through this exploratory search align with the hand-written rules, validating their optimality to a certain degree. During testing, we adjust $p_{\text{sample}}=0.5$ to ensure a more efficient search and a higher success rate in task completion.
    
    \item \textbf{$\text{MCTS}_{\text{TG}}$ and $\text{MCTS}_{\text{RG}}$.}
    We develop \textit{$\text{MCTS}_{\text{TG}}$} and \textit{$\text{MCTS}_{\text{RG}}$} to replicate the baselines used in Watch-and-Help~\cite{puig2020watch}. In the original work, a hierarchical planner was implemented, using MCTS for high-level planning and regression planning (RP) for low-level planning. Since our task environment does not require low-level planning, we use the \textit{MCTS-heuristic} model as a counterpart of their planner in this specific environment.
    To provide comparison, we implemented two additional baselines: \textit{$\text{MCTS}_{\text{TG}}$}, which has knowledge of the main agent's true goal, and \textit{$\text{MCTS}_{\text{RG}}$}, which follows a random goal. As shown in the results table, the performance of \textit{$\text{MCTS}_{\text{RG}}$} is similar to the \textit{Random} agent, with minimal ability to provide assistance and complete tasks. This highlights the importance of a smart agent that can infer the main agent's goal.
    \textit{$\text{MCTS}_{\text{TG}}$}, knowing the true goal of the main agent, achieves the best performance across all metrics. However, while it significantly surpasses other models in other metrics, completing the task better and faster, its \textit{HN} is relatively low, indicating the limitation of its simple take-over helping strategy.

\end{itemize}

\end{document}